\documentclass[runningheads]{llncs}
\usepackage[T1]{fontenc}
\usepackage{graphicx,verbatim}
\usepackage{booktabs}
\usepackage{multirow}
\usepackage{colortbl}
\usepackage{arydshln}
\usepackage{xcolor}
\definecolor{vesselsim}{RGB}{245,225,255}   
\definecolor{finetune}{RGB}{255,250,205}    
\definecolor{DarkGreen}{RGB}{0,100,0}

\begin{document}

\title{VesselSim: learning 3D blood vessel segmentation without expert annotations}

\author{Erin Rainville\inst{1}\thanks{Corresponding author.} \and
Melissa Ananian\inst{1}\and
Tristan Mirolla\inst{1}\ \and
Hassan Rivaz\inst{2}\and
Yiming Xiao\inst{1}}


\authorrunning{Erin Rainville et al.}

\institute{Department of Computer Science and Software Engineering, Concordia University, Montreal, Canada \\
    \email{e\_ainvil@live.concordia.ca}}

\institute{Department of Electrical and Computer Engineering, Concordia University, Montreal, Canada \\ 
\and
Department of Computer Science and Software Engineering, Concordia University, Montreal, Canada
\\
\email{e\_ainvil@live.concordia.ca}}
  
\maketitle              
\begin{abstract}
Blood vessel segmentation is a core task in medical image analysis for the care of vascular diseases and surgical planning, yet the challenges of providing expert vascular annotations pose a major obstacle for the progress of related deep learning techniques. To address this, we propose \texttt{VesselSim}, a two-stage framework for universal 3D blood vessel segmentation that eliminates the need for real annotated data during training. First, we introduce a stochastic, geometry-driven vascular simulation framework that models recursive branching, curvature-controlled growth, and collision-aware topology, followed by domain-randomized intensity synthesis to generate 16,500 anatomically plausible 3D angiographic volumes. Second, a 3D U-Net is trained solely on this synthetic data. To bridge the domain gap from synthetic to real images at inference time, we introduce a test-time adaptation strategy via a self-supervised mask reconstruction decoder, enabling adaptation to unseen clinical scans without prior domain knowledge. We evaluate \texttt{VesselSim} in a zero-shot setting on multiple real-world datasets spanning MR and CT across several anatomical regions, including the brain and kidneys. Despite being trained exclusively on synthetic data, \texttt{VesselSim} achieves performance competitive with state-of-the-art vascular segmentation foundation models. These findings suggest that learning vessel geometry from synthetic tubular structures is effective for robust cross-domain generalization, substantially reducing the reliance on acquired medical imaging data and more importantly, expert annotations.

\keywords{Vascular segmentation \and synthetic data \and test-time training.}

\end{abstract}

\section{Introduction}
Blood vessel segmentation from 3D medical scans plays a critical role in various clinical applications, including diagnosing vascular disorders \cite{MOCCIA201871}, planning surgical interventions \cite{Beriault2015}, and studying disease-related perfusion abnormality as biomarkers \cite{haast2024insights}. With delicate geometry, soft contrast, complex anatomical context, and widespread spatial distribution, manual labeling of vasculature is often much more challenging and time-consuming than other body organs. Largely due to this, publicly labeled 3D vascular datasets are scarce, greatly hindering the development of deep-learning (DL)-based vascular segmentation methods.

To address this, recent advances in data-efficient 3D vessel segmentation focus on reducing annotation demand and improving domain generalization. Few-shot learning approaches tailored to vascular structures have shown promise. VesselShot \cite{aktar2023vesselshot} explores few-shot cerebrovascular segmentation by leveraging knowledge from few labeled support samples. FSVS-Net \cite{YANG2025103281} introduces a semi- supervised few-shot framework that uses feature distillation and bidirectional weighted fusion to propagate sparse annotated slices to unlabeled ones for hepatic, pulmonary, and renal vessel segmentation. Later, biomedical foundation models have also advanced data efficiency for the task. Notably, VesselFM \cite{wittmann2025vesselfm} is a universal foundation model specifically trained for 3D blood vessel segmentation using a combination of domain-randomized synthetic images, flow-matching generated samples, and real annotated data; enabling robust zero- and few-shot performance across diverse scans. Besides algorithm development, to further mitigate the dependency on expert annotations, a few rule- and DL-based approaches for vessel simulation \cite{feldman2025vesselgpt,feldman2023vesselvae,guo2025vesseldiffusion,hamarneh2010vascusynth,sweeney2024unsupervised} have also been proposed, but they are often domain-specific (with reliance on training data) or lack of typical curvy geometries found in vasculatures, potentially limiting their generalization capacity. In addition to training samples, domain shift poses a challenge for the robustness of vessel segmentation algorithms. More recently, test-time adaptation (TTA) methods \cite{Yu2025ALS} that use unlabeled data to efficiently align model features or regularize outputs at inference time have shown great power in medical image segmentation. Yet, despite early investigations with 2D scans \cite{gu2025test,zhou2025topotta}, \textit{very few have explored TTA in the realm of 3D blood vessel segmentation}.

As existing data-efficient 3D vessel segmentation methods still rely on real vascular annotations and lack robustness to domain shift, we propose \texttt{VesselSim}, a new blood vessel segmentation DL model trained on data synthesized by a probabilistic rules-based simulation for more natural looking geometries, without manual segmentation ground truths. Our two-stage framework overcomes the aforementioned limitations by combining topology-constrained synthetic pretraining with ground-truth-free test-time adaptation, enabling structurally consistent segmentation and improved generalization from synthetic to real clinical volumes. \textit{Our major contributions are three-fold}: \textbf{First}, we proposed a new stochastic technique to simulate artificial 3D angiography scans at different scales for DL model training, and release the code and dataset to the community at \textcolor{blue}{\url{\textit{https://healthx-lab.github.io/VesselSim}}}. \textbf{Second}, we investigate a novel test-time adaptation technique based on  mask reconstruction to efficiently bridge domain shifts between synthetic and real clinical data without needing voxel-wise annotations. \textbf{Lastly}, we compare the proposed method against state-of-the-art foundation models across multiple 3D vascular segmentation tasks. 

\begin{figure}
\includegraphics[width=\textwidth]{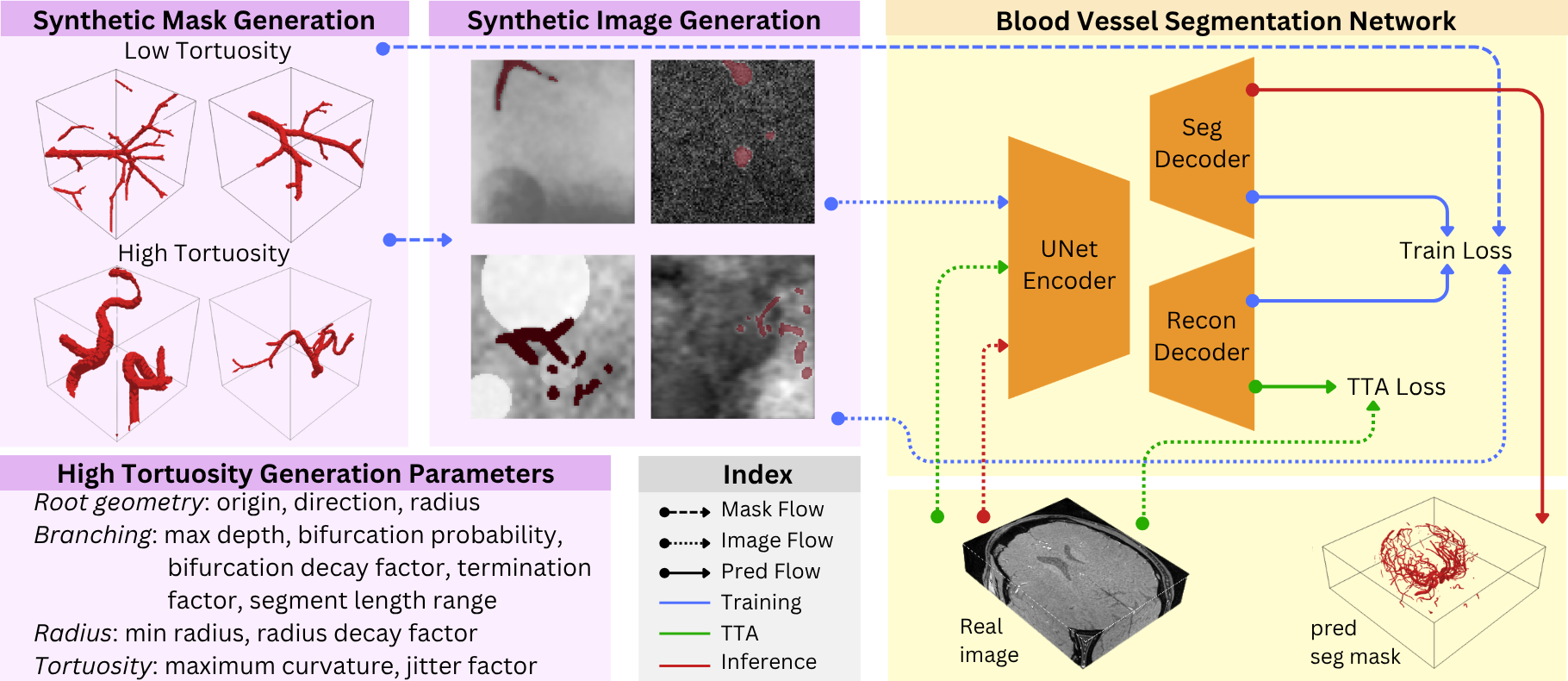}
\caption{VesselSim framework. Left: Synthetic vascular masks are generated with controllable geometric parameters. Masks are converted into angiography-like volumes via domain-randomization. Right: A 3D U-Net trained on the synthetic data. At inference, the self-supervised reconstruction decoder enables test-time adaptation to real scans.} \label{fig1}
\end{figure}
\noindent

\section{Methods and Materials}
The overall pipeline of the proposed angiography simulation method and deep learning framework with test-time adaptation for blood vessel segmentation are summarized in Fig.~\ref{fig1}.

\subsection{Artificial vascular image generation}
Existing vascular simulation methods vary in realism and data dependence. Classical rule-based approaches \cite{hamarneh2010vascusynth,valverde2013lsystem} rely on recursive branching or cost-minimization principles with the possibility of generating large, complex vascular networks, but typically result in stick-like or piecewise-linear segments that insufficiently represent the tortuous/sinuous geometry of certain vasculature (e.g., in the brain). In contrast, deep generative models, such as VesselGPT \cite{feldman2025vesselgpt} and VesselVAE \cite{feldman2023vesselvae} achieve more realistic cerebrovascular morphologies, but require large annotated clinical datasets and are restricted to short local vessel fragments. 

To enable multi-scale, data-independent vascular synthesis, we propose a procedural stochastic branching framework. Extending from \cite{hamarneh2010vascusynth}, we incorporate a biased random walk into a stochastic recursive branching system to control segment trajectories, enabling curvature and directional persistence. Segment length, rotation axis, and rotation magnitude are sampled from defined probabilistic ranges informed by the literature \cite{wang2013design,jessen2023branching}, and growth proceeds in a \textit{breadth-first manner} within a $160\times160\times160 $ voxel volume to ensure uniform spatial density. Collision avoidance is enforced via multi-attempt path validation, terminating branches when no feasible path exists. Additionally, branching probability decays with depth while radii decrease following physiological scaling consistent with Murray’s law \cite{murray1926physiological}. \textit{To generate the final angiography volume} (see Fig.~\ref{fig1}), we extract up to 50 random $96\times96\times96 $-voxel sub-volumes from each main one that pass an automatic quality control based on criteria of flow continuity, vessel integrity (no small "floating islands"), and a 5\% minimum vessel occupancy threshold. Then, we augment random 3D geometry shapes as background of the vessel blocks and Gaussian noise to simulate MRI and CT angiographies, following VesselFM's domain randomization scheme \cite{wittmann2025vesselfm}. Notably, we intentionally added ellipsoidal shell shapes to simulate skulls to improve performance. For the full training dataset, we combine 5000 low-tortuosity vessel samples with VascuSynth \cite{hamarneh2010vascusynth}, 5000 high-tortuosity ones with our method, 2500 skull samples, and 2500 background samples to reduce false positive rates. An additional 500 low-tortuosity, 500 high-tortuosity, 250 skull-injected and 250 background samples are separately generated as the validation set to prevent data leaking between training and validation.

\subsection{Learning with synthetic vascular data}

\texttt{VesselSim} is based on a 3D UNet architecture \cite{monai} trained on synthetic data from our vessel generation framework in an end-to-end manner. To encourage accurate segmentation of thin, branching vascular structures, we employ a composite objective. The primary segmentation loss consists of Cross-Entropy ($\mathcal{L}_{CE}$) and Dice loss ($\mathcal{L}_{Dice}$), balancing voxel-wise classification with volumetric overlap. To further penalize topological discontinuities and centerline fragmentation, we integrate a centerline boundary Dice loss ($\mathcal{L}_{cbDice}$) \cite{shi2024centerline}. This term leverages soft-skeletonization to enforce alignment between predicted and ground-truth vessel centerlines and boundaries, providing an explicit inductive bias toward tubular geometry and connectivity.

To facilitate domain adaptation from synthetic to real images, we extend the segmentation network with a self-supervised auxiliary mask reconstruction task. We attach an auxiliary reconstruction decoder to the shared encoder bottleneck to perform in-painting. During training, input volumes are corrupted by randomly masking multiple cubical regions (ranging from $2\times2\times2$ to $16\times16\times16$ voxels) with a constant intensity value  of $-1$ \cite{pathak2016inpainting,devries2017cutout}. The network is jointly optimized for segmentation and reconstruction task ($\mathcal{L}_{rec}$) so that the axillary task can provide better alignment in TTA. The total training objective is:

\begin{equation}
\mathcal{L}_{total} = \alpha\mathcal{L}_{CE} + \beta\mathcal{L}_{Dice} + \gamma\mathcal{L}_{cbDice} + \lambda\mathcal{L}_{rec}
\end{equation}
where weights are empirically set to $\alpha=0.1, \beta=0.9, \gamma=0.2,$ and $\lambda=0.1$. The model is trained exclusively on the generated dataset of 16,500 volumetric patches, using a single NVIDIA H100 GPU with a batch size of 16. Optimization is performed with an initial learning rate of $10^{-4}$, followed by linear decay to $10^{-5}$ until convergence. All input image intensities are normalized to [0,1].

\subsection{Test-time adaptation for vascular segmentation}

We leverage the reconstruction decoder as a self-supervised test-time adaptation mechanism to mitigate domain shift between synthetic training data and real-world clinical images, which can differ in contrast, noise characteristics, intensity scaling, acquisition artifacts, and anatomical features. The reconstruction objective provides an unsupervised alignment signal that encourages the encoder to adjust its internal representations to the appearance characteristics of each target volume, without requiring any expert annotations.

At inference time, we randomly sample 24 patches of $96\times96\times96$ voxels from 4 scans of the unseen test set and apply the same masking strategy used during training. The reconstruction loss $\mathcal{L}_{rec}$ is computed between the reconstructed outputs and the original patches. After a single adaptation epoch with a learning rate of $10^{-4}$, only the instance normalization layer parameters within the encoder are updated while all other weights remain frozen. This targeted update selectively allows the network to recalibrate its activation distributions to the target domain efficiently while preserving the learned structural vascular priors. The final segmentation is obtained by forwarding the original, uncorrupted test volume through the newly adapted model. As a final refinement step, small disconnected components below a predefined volume threshold are removed to suppress spurious non-vascular predictions.

\subsection{Real vascular data and comparison implementation}
To evaluate generalization beyond synthetic data, we perform inference on two publicly available medical image repositories, \textit{HiP-CT} and \textit{TopCoW} (3 distinct datasets in total), thus allowing us to assess robustness across organ systems (kidney and brain) and across imaging contrasts (HiP-CT, CTA and MRA).

The \textbf{\textit{HiP-CT}} database \cite{jain2023sennet} consists of three ultra–high-resolution 3D human kidney volumes acquired using Hierarchical Phase-Contrast Tomography. The original volumes have an average dimension of $1465\times1494\times1732$ voxels with an isotropic resolution from $5um$ to $50um$. Due to their large size, volumes are partitioned into $500^3$ voxel sub-volumes for computational feasibility. These sub-volumes are used consistently across all baseline models and experiments.

The \textbf{\textit{TopCoW}} dataset \cite{topcowchallenge} includes 125 computed tomography angiographies (CTA) and 125 magnetic resonance angiographies (MRA) of the human brain. These modalities differ substantially in contrast mechanisms and the image characteristics. While the CTA volumes have an average resolution of  $0.45\times0.45\times0.7$ $mm^{3}$, the MRA volumes average $0.3\times0.3\times0.6$ $mm^{3}$. All scans are resampled to an common resolution of $0.4\times0.4\times0.4\,mm^3$. For \textit{\textbf{TopCoW CTA}}, manual annotations are provided for the Circle of Willis. Volumes are cropped to the bounding box of this region to avoid penalizing predictions outside the territory. For \textbf{\textit{TopCoW MRA}}, we use the VesselVerse extension \cite{vesselverse}, which expands the TopCoW MRA annotations to for the full cerebrovasculature.

To contextualize the performance of \texttt{VesselSim} on these datasets, we compare against three recent state-of-the-art (SOTA) medical image segmentation foundation models evaluated on the same test sets. UniverSeg \cite{butoi2023universeg} is a general-purpose prompt-based medical segmentation model whose capability for vascular segmentation has been previously demonstrated. It is trained on 53 heterogeneous medical segmentation datasets and performs task adaptation through a support-set prompting mechanism. For each target dataset, we provide four randomly selected volumes as a support set to condition the model on the segmentation task. We additionally evaluate SAM-Med3D \cite{wang2024sammed}, a 3D extension of the Segment Anything Model tailored for volumetric medical imaging. SAM-Med3D leverages large-scale pretraining and prompt-based mask generation to enable adaptable segmentation across anatomical structures. VesselFM \cite{wittmann2025vesselfm} is a foundation model specialized for vascular segmentation, pretrained across 19 vessel datasets and applied here in a zero-shot manner without additional adaptation. Notably, both TopCoW and HiP-CT are included in the VesselFM training corpus, potentially conferring an inherent advantage in this evaluation. In contrast to all foundation baselines, \texttt{VesselSim} is trained exclusively on synthetic vascular data and has no exposure to real-world clinical images during training.

As the foundation models are pretrained on large-scale real-world datasets, we additionally evaluate whether \texttt{VesselSim} can benefit from limited exposure to real target-domain data. For each dataset, we perform a small-sample model finetuning of the procedure using the same four support volumes used for UniverSeg. From each volume, we randomly sample 24 sets of $96\times96\times96$ patches, resulting in 96 training patches per dataset. The model is then finetuned for a single epoch using a learning rate of $10^{-5}$, updating all network weights. This setting simulates a realistic low-annotation regime, where only a small number of labeled scans are available.

\section{Results}

\begin{table*}[t]
\caption{Comparison of segmentation performance across datasets. \texttt{VesselSim} is our final model with loss \textit{CE + Dice + cbDice + Reconstruction TTA}. The best results for each dataset (Dice \%, clDice \%) are in bold, the second best results are underlined. The last row denotes the small-sample finetuning upper bound of \texttt{VesselSim} and is not included in best and second best highlights.}
\centering
\small
\setlength{\tabcolsep}{2pt}
\renewcommand{\arraystretch}{1.1}
\begin{tabular}{lcc cc cc}
\toprule
\multirow{2}{*}{Models} & \multicolumn{2}{c}{TopCoW CTA} & \multicolumn{2}{c}{TopCoW MRA} & \multicolumn{2}{c}{HiP-CT} \\
\cmidrule(lr){2-3}
\cmidrule(lr){4-5}
\cmidrule(lr){6-7}
& Dice & clDice  & Dice  & clDice  & Dice & clDice  \\
\midrule
VesselFM 
& \textbf{52.9$\pm$9.5} 
& 58.9$\pm$10.2 
& 48.3$\pm$8.1 
& 46.9$\pm$9.1 
& \underline{35.1$\pm$24.5} 
& 25.6$\pm$19.5 \\

SAM-Med3D 
& 6.6$\pm$4.4 
& 9.1$\pm$6.0 
& 3.1$\pm$2.6  
& 2.6$\pm$2.2 
& 8.0$\pm$21.6 
& 3.1$\pm$8.9  \\

UniverSeg 
& 19.7$\pm$8.6 
& 29.4$\pm$12.3 
& 48.2$\pm$5.1  
& 47.0$\pm$6.1 
& 4.7$\pm$21.1 
& 0.0$\pm$0.0  \\

\hdashline

Base UNet 
& 42.7$\pm$12.2 
& 60.6$\pm$12.7 
& 70.1$\pm$3.9 
& 73.1$\pm$5.1 
& 34.3$\pm$23.0  
& 23.0$\pm$42.7 \\

+ cbDice 
& 46.8$\pm$12.9 
& \underline{62.9$\pm$13.6} 
& \textbf{72.6$\pm$3.5}  
& \underline{75.7$\pm$4.4} 
& 31.1$\pm$23.7  
& 41.2$\pm$25.1 \\

+ recon 
& 43.1$\pm$14.1 
& 60.6$\pm$13.3 
& \underline{71.9$\pm$3.9} 
& 74.6$\pm$5.2 
& \textbf{36.0$\pm$22.7}  
& \underline{41.3$\pm$23.1}  \\

\rowcolor{vesselsim}
VesselSim 
& \underline{48.7$\pm$11.9} 
& \textbf{64.0$\pm$13.0} 
& 71.8$\pm$3.3
& \textbf{76.2$\pm$4.1} 
& 31.6$\pm$21.5
& \textbf{42.3$\pm$23.8}  \\

\rowcolor{finetune}
+ finetuned   
& 59.5$\pm$7.7 
& 70.4$\pm$7.3 
& 77.3$\pm$3.3 
& 83.2$\pm$3.6 
& 46.8$\pm$33.2 
& 50.7$\pm$27.8  \\

\bottomrule
\end{tabular}
\label{tab:main_results}
\end{table*}

Table~\ref{tab:main_results} summarizes the comparison between \texttt{VesselSim} variants (for ablation study) and foundation model baselines across all datasets in terms of Dice and clDice for volumetric accuracy and topological consistency of tubular structures. Overall, the zero-shot \texttt{VesselSim} trained solely on our synthetic dataset achieves competitive performance relative to foundation models trained on substantially larger real clinical datasets. Notably, it generalizes well across both neurovascular (CTA/MRA) and renal CT domains, outperforming or matching the baselines. This demonstrates that synthetic-only training can transfer effectively to real clinical data across organs and imaging modalities.

HiP-CT represents an extreme domain shift due to its ultra-high-resolution imaging characteristics. In this setting, UniverSeg and SAM-Med3D fail to detect vessels reliably, resulting in near-zero Dice and clDice scores. VesselFM was pretrained on HiP-CT so it performs well. In contrast, \texttt{VesselSim} achieves competitive performance, demonstrating that geometry-driven synthetic training enables robust generalization even under severe appearance shifts.

The ablation study comparing different \texttt{VesselSim} variants reveals the contribution of different key training components. Starting from the based 3D UNet (CE + Dice), incorporating cbDice primarily improves topological consistency, reflected in systematic gains in clDice across modalities. In contrast, the reconstruction auxiliary task yields larger improvements in cross-domain CT performance, suggesting enhanced robustness to appearance shifts. With reconstruction-based TTA for real scans produces the strongest overall model, consistently improving structural preservation while maintaining or improving volumetric overlap. These complementary effects indicate that cbDice promotes geometric continuity of thin tubular structures, whereas reconstruction-based adaptation mitigates modality-specific intensity differences. Finally, small-sample fine-tuning can further boosts performance across all datasets, particularly in the cross-organ CT setting. This indicates that synthetic pretraining provides a strong prior that can be effectively refined with minimal real supervision.

\begin{figure}
\centering
\includegraphics[width=\textwidth]{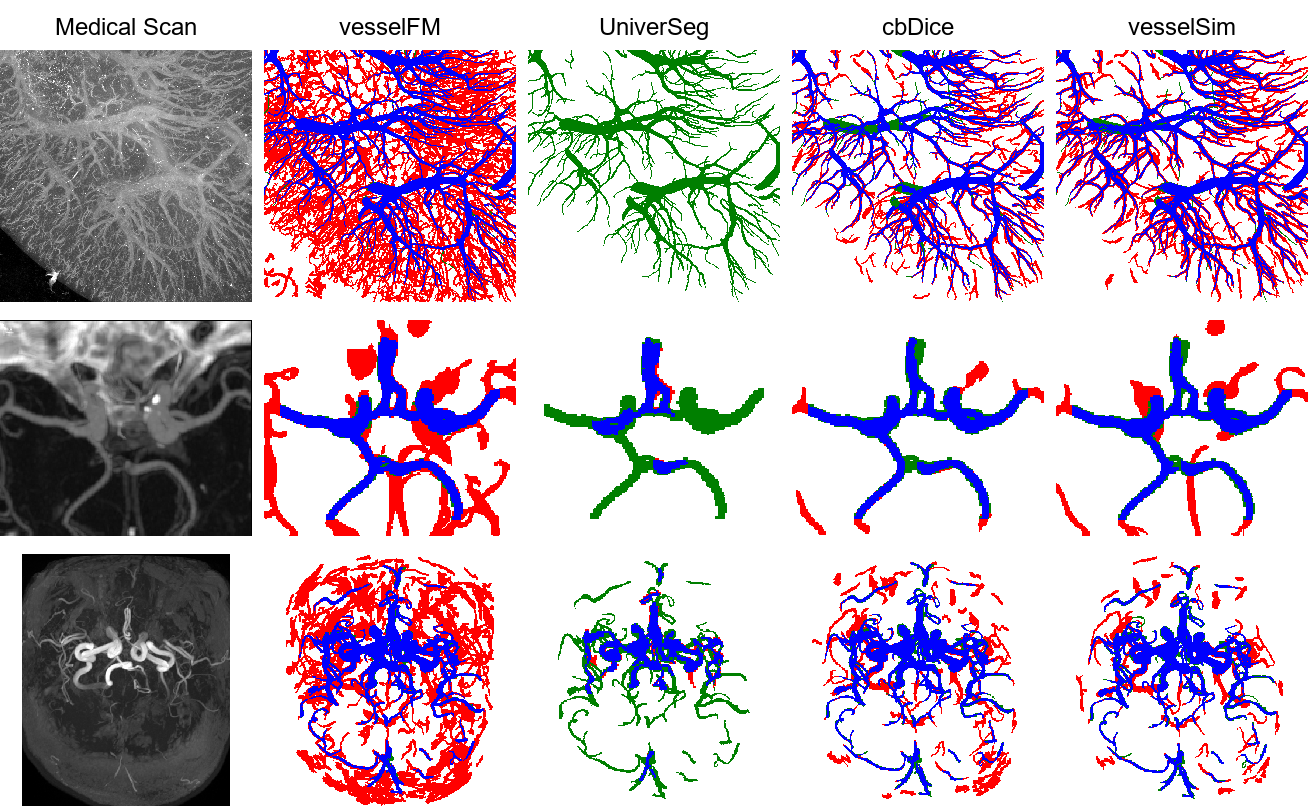}
\caption{Maximum intensity projections of the input scan and different model predictions. From top to bottom: HiP-CT,TopCoW CTA,TopCoW MRA. For the predictions, True positives are shown in {\color{blue}blue}, false positives in {\color{red}red}, and false negatives in {\color{DarkGreen}green}.  
} \label{fig2}
\end{figure}

\section{Discussion}
The proposed \texttt{VesselSim} only relies on training from 15,000 synthetic vascular 3D patches \underline{without any expert annotation}, in contrast to baseline foundation models that are trained on substantially larger real-data corpora (e.g., specialized VesselFM trained on 115,000 real angiography patches in addition to extensive domain-randomized data simulated from expert annotations). Yet, \texttt{VesselSim} matches or surpasses VesselFM, SAM-Med3D, and UniverSeg on multiple datasets. A few design choices contribute to the performance and training efficiency. \textbf{First}, the proposed enhanced stochastic vessel generation algorithm in addition to domain randomization helps produce more natural looking vascular patches, which are beneficial for learning strong prior to compensate for the absence of large-scale real supervision. Also, due to the challenges in manual annotation of blood vessels, expert labels can often miss true targets while synthetic data offers more confident ground truths. Notably, with the additional skull simulation, the mean Dice score was improved from 0.61 to 0.72 before TTA for the TopCoW MRA dataset. \textbf{Second}, with the addition of mask reconstruction auxiliary task and cbDice loss, the training emphasizes on learning tubular structural features. Notably, improvement in clDice across modalities further supports the hypothesis that the model learns geometry-driven representations, considering VesselFM does not include additional losses designed for tubular features. \textbf{Lastly}, by including the auxiliary task intended for TTA during main model training, it helps guide domain adaptation for segmentation \cite{colussi2025rec}. 

\noindent
For TTA, we focused on adapting the instance normalization layers of the 3D UNet than the full encoder. This approach intends to help mitigate ``weight drift" under small sample adaption, and offers better computational efficiency for real-time deployment. Notably, for TopCoW MRA, adapting the full encoder reduces the mean Dice and clDice scores by 4.5\% and 8.3\%, respectively.   

\noindent
When inspecting the results more closely, we find that remaining zero-shot errors are primarily associated with larger-caliber vessels (Fig.~\ref{fig2}), which are underrepresented in the current generation process. Future work will expand scale diversity within our controllable parameter framework (e.g., increasing initial branch radius) to better model multi-scale vascular hierarchies. Overall, these findings highlight the feasibility of greatly reducing dependence on large-scale real annotations through structured synthetic data generation while maintaining competitive cross-domain generalization.

\section{Conclusion}
To address the critical challenge of limited annotated datasets for advancing deep learning-based 3D blood vessel segmentation, we proposed \texttt{VesselSim}, a novel two-stage learning framework based on both simulated and real data using unsupervised test-time adaptation. Without needing a single manual segmentation ground truths, the proposed method demonstrated excellent performance across modalities and tasks in comparison with state-of-the-art methods, offering a new direction for data-efficient vascular segmentation.  

\begin{credits}
\subsubsection{\ackname} We acknowledge the support of the Natural Sciences and Engineering Research Council of Canada (NSERC) and the Fonds de recherche du Québec – Nature et technologies (FRQNT). 

\subsubsection{\discintname}
The authors have no competing interests.

\end{credits}

%
%
%
 \bibliographystyle{splncs04}
 \bibliography{mybib}

\end{document}